\documentclass[letterpaper, 10 pt, conference]{ieeeconf}  

\IEEEoverridecommandlockouts                              

\usepackage{graphics} 
\usepackage{epsfig} 
\usepackage{mathptmx} 
\usepackage{times} 
\usepackage{amsmath} 
\usepackage{amssymb}  
\usepackage{adjustbox}
\usepackage{censor}

\title{Inject, Encode, Navigate: Feature-Wise Time Injection for Rectified Flow with Bidirectional Feature Encoding}

\author{Xiaogang Wang$^{1}$ Debasish Mishra$^{1}$ Jia Hao Toh$^{2}$ Wangsheng Liu$^{2}$ Yash Shah$^{1}$%
\thanks{$^{1}$Xiaogang Wang, Debasish Mishra and Yash Shah are with Amazon Web Services.
        {\tt\small \{xiaogawa, misdebas, syash\}@amazon.com}}%
\thanks{$^{2}$Jia Hao Toh and Wangsheng Liu are with Certis Group.
        {\tt\small \{EvinJH\_TOH, wangsheng\_liu\}@certisgroup.com}}%
}

\begin{document}

\maketitle
\thispagestyle{empty}
\pagestyle{empty}

\begin{abstract}

Vision-model-based robotic navigation has gained increasing popularity, yet existing approaches still struggle to deliver effective and efficient navigation performance, \textit{e.g.}, diffusion-based policies that require costly iterative denoising. In this work, we propose three improvements to bridge these gaps.
First, we adopt a rectified flow framework to reduce the required inference steps to one while achieving comparable or even superior performance. 
Second, we leverage a bidirectional Mamba-based visual encoder that employs selective state-space modeling to more effectively capture bidirectional temporal dependencies across observation frames, even with longer input sequences. 
Most importantly, we find that na\"{i}vely combining these two components to balance accuracy with inference speed yields no further performance gains or even produces inferior results. 
Upon extensive analysis, we realized that the root cause lies in the fact that the
Hydra-derived temporal context is shared unchanged across all flow
timesteps, and timestep information is introduced only within the
velocity network. This weak coupling limits the model's ability to
adaptively exploit temporal visual features at different flow states.
To address this, we propose a feature-wise timestep conditioning design that  
explicitly conditions the visual embedding on the flow timestep, enabling noise-level-aware trajectory generation,
thereby restoring a meaningful gradient pathway and enabling coherent interaction between the rectified flow and the temporal encoder.
Extensive quantitative and qualitative experiments across multiple datasets demonstrate that our approach consistently outperforms both baseline and competitive methods.

\end{abstract}

\section{INTRODUCTION}

Autonomous visual navigation is the ability of a robot to perceive its surroundings through onboard sensors and navigate toward a designated goal without relying on pre-built maps, which is a cornerstone capability for deploying intelligent agents in the real world. The importance of this task spans a broad spectrum of applications. In security and surveillance, mobile robots need to patrol large facilities, detect anomalies, and autonomously navigate to points of interest in dynamic and unstructured environments. In search and rescue, agents are required to traverse disaster-stricken terrains where GPS signals are unavailable and pre-existing maps are rendered obsolete. Moreover, warehouse logistics, last-mile delivery, agricultural inspection, and assistive robotics for the elderly all demand robust, real-time visual navigation that generalizes across diverse and previously unseen environments. 

\begin{figure}[h!]
    \centering
    \includegraphics[width=0.48\textwidth]{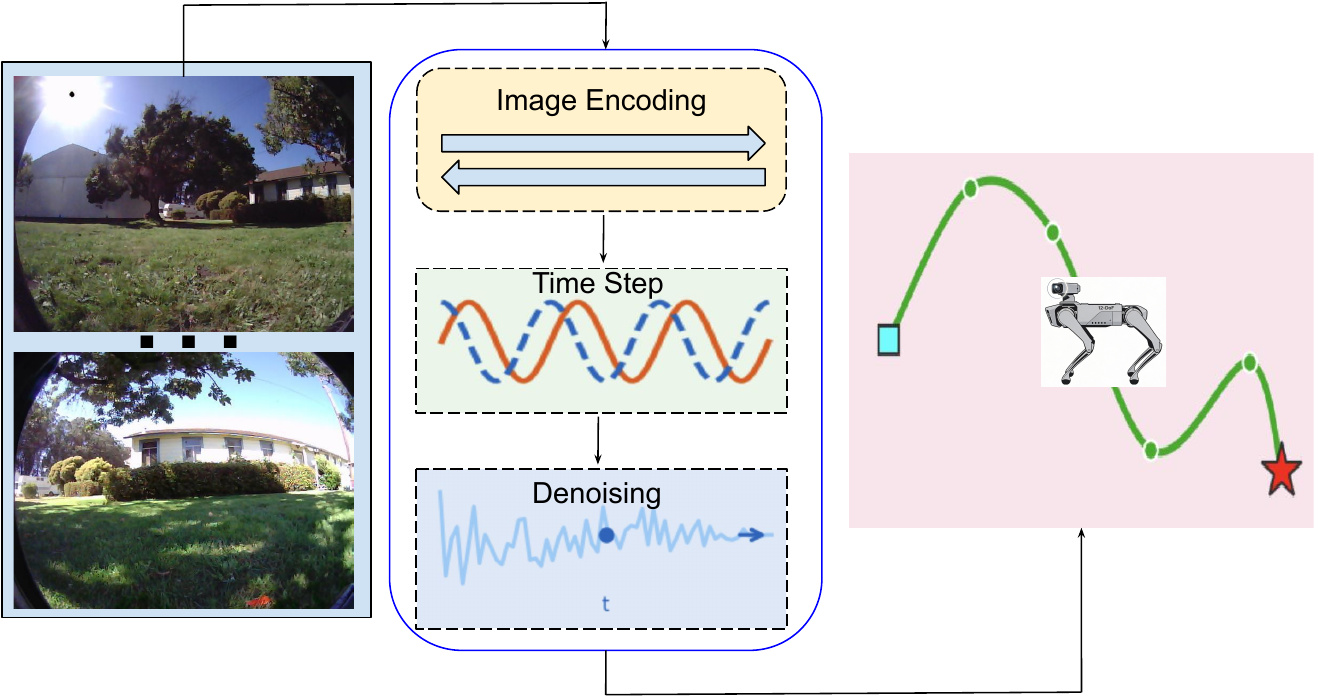}
    \vspace{-8mm}
    \caption{Our navigation model takes visual observations and generates denoised action trajectories, enabling a mobile robot to navigate toward the target in real-world environments.}
    \vspace{-7mm}
    \label{fig:teaser}
\end{figure}

The recent emergence of large vision navigation models~\cite{shah2023vint,sridhar2024nomad,cao2024mamba,peng2025lovon,du2025vlnav,zhang2025apexnav} has opened a promising avenue for this task, enabling robots to ground visual goals into actionable navigation policies. These advances have brought the community closer than ever to deploying general-purpose navigation agents that can operate reliably outside controlled laboratory settings.
Early learning-based approaches such as ViNT~\cite{shah2023vint} introduced the idea of training a general-purpose navigation policy across diverse robotic platforms and environments, demonstrating impressive cross-embodiment transfer. More recently, NoMaD~\cite{sridhar2024nomad} extended this line of work by incorporating a diffusion-based action generation policy, leveraging the expressiveness of Denoising Diffusion Probabilistic Models (DDPM)~\cite{ho2020denoising} to produce multi-modal waypoint distributions and achieving strong performance in topological navigation tasks. However, diffusion-based policies inherently require multiple iterative denoising steps at inference time which introduces additional computational latency. Furthermore, existing methods mostly rely on transformer-based~\cite{vaswani2017attention} or convolutional encoders to process multi-frame visual observations. While effective for short observation histories, these encoders struggle to capture long-range temporal dependencies as the number of input frames grows, due to the quadratic complexity of self-attention~\cite{vaswani2017attention} or the limited receptive fields of convolutional architectures. This results in less accurate waypoint predictions, especially in scenarios that require reasoning over extended navigation histories to resolve ambiguities in the environment. 

To address these challenges, we propose three synergistic improvements built upon state-of-the-art diffusion-based framework NoMad~\cite{sridhar2024nomad}.
Firstly, to speed up the inference steps, we propose to replace the conventional DDPM denoising process~\cite{ho2020denoising} with a Rectified Flow Matching~\cite{liu2023flow, lipman2023flow} formulation that reformulates waypoint prediction as learning straight-line trajectories in data space. This principled reformulation reduces the required inference steps from ten to just one via a single Euler integration step, while achieving on-par or even superior waypoint prediction accuracy compared to the DDPM strategy~\cite{ho2020denoising}.
Secondly, to strengthen temporal context modeling across multi-frame observations, we propose to adopt the bidirectional Mamba-based model Hydra~\cite{gu2024mamba} to process the input observation images inspired by~\cite{cao2024mamba}. Unlike the transformer's quadratic self-attention, Mamba's linear recurrence mechanism captures long-range temporal dependencies far more effectively as the context window grows, yielding richer goal-conditioned visual representations particularly when processing large numbers of input frames. In our experiments, we are able to consume up to ten frames.
Thirdly and most critically, we observe that naively combining the above
two improvements does not provide cumulative gains and can even degrade
performance. We attribute this behavior to insufficient interaction
between the flow state and the temporally enriched visual representation.
Specifically, the Hydra encoder produces a context vector that remains
unchanged across flow timesteps, while the timestep is provided only to
the downstream velocity network. As a result, the same visual feature
representation is used throughout flow-matching training, regardless of
the current interpolation state.
To address this, we introduce a lightweight feature-wise conditioning mechanism that injects the flow timestep into the visual conditioning signal before it enters the denoising network, restoring noise-level awareness so that the two components work in synergy. 
Extensive quantitative and qualitative experiments across multiple robot navigation benchmarks demonstrate that our unified framework achieves superior waypoint prediction accuracy and navigation success rates compared to existing works.

Last but not least, we further extend our proposed architecture with a detection-guided mechanism to tackle a two-phase security inspection task using the Deep Robotics X30 Pro quadruped robot: first navigating the robot toward a target object, then executing a circular motion around it for detailed inspection. By incorporating a lightweight object detection head~\cite{ren2016faster} into our framework, the robot can approach a specific target without relying on a goal frame image, using the bounding box area as an additional indicator for autonomous stopping. 
This mechanism builds naturally upon our flow matching policy and the learned model generates the overall navigation and circling trajectories, while the detection head enhances target-aware approaching.
An illustration of our navigation task is shown in Figure~\ref{fig:teaser}.

Our main contributions can be summarized as follows:
\begin{itemize}
    \item We carefully incorporate rectified flow matching into the diffusion-based framework~\cite{sridhar2024nomad} to reduce inference steps from teens to one while also improving waypoint prediction accuracy.
    \item We propose to replace the transformer-based temporal encoder with a bidirectional Mamba state-space model Hydra, which more effectively captures long-range temporal dependencies across multi-frame observations with linear computational complexity.
    \item We introduce a feature-wise timestep conditioning mechanism that directly modulates the temporally enriched visual context according to the current flow state, enabling stronger interaction between temporal perception and trajectory generation.
    \item We conduct extensive experiments on multiple datasets, demonstrating that our unified framework achieves state-of-the-art performances.
    \item We extend our framework with a detection-guided mechanism for a two-phase security inspection task, where the object detector enables goal-image-free target approaching.
\end{itemize}

\section{RELATED WORKS}

\subsection{Vision-Based Robot Navigation}

Vision-based navigation has evolved from classical map-based planning to modern learning-based methods. 
Recent works have focused on general-purpose, transferable navigation policies including ViNT~\cite{shah2023vint} learned a universal policy from heterogeneous multi-robot datasets with zero-shot transfer, GNM~\cite{shah2022gnm} established the foundation model paradigm for goal-conditioned navigation, and NoMaD~\cite{sridhar2024nomad} integrated diffusion-based action generation to capture multi-modal action distributions for state-of-the-art topological navigation. Despite this progress, these methods still face limitations in larger temporal context modeling, and diffusion-based method needs several inference steps to get reasonable results.

\subsection{Diffusion Models for Robot Control}

Diffusion models have achieved remarkable success in image generation and have been increasingly adopted for robot control. Diffusion Policy~\cite{chi2025diffusion} pioneered applying DDPM~\cite{ho2020denoising} to visuomotor policy learning, demonstrating that iterative denoising effectively captures multi-modal action distributions in manipulation tasks. This approach has since been extended to navigation~\cite{sridhar2024nomad} and other robotic domains. However, diffusion-based policies suffer from high inference latency due to multiple sequential denoising steps. While DDIM~\cite{song2020denoising} has attempted to reduce sampling steps through non-Markovian processes, our work takes a fundamentally different approach by reformulating action generation using flow matching~\cite{liu2023flow}, which inherently supports single-step inference without requiring distillation or additional training stages.

\subsection{Flow Matching and Rectified Flow}

Flow matching~\cite{lipman2023flow, liu2023flow} has emerged as an elegant alternative to diffusion models, learning a deterministic velocity field that transports samples from a simple prior to the data distribution, rather than reversing a stochastic noising process. Rectified Flow~\cite{liu2023flow} further simplifies this by defining straight-line interpolation paths, enabling accurate single-step Euler integration. While flow matching has shown competitive or superior performance in image~\cite{esser2024scaling} and video generation with significant sampling efficiency gains, its application in robotics remains underexplored. To our knowledge, no prior work has systematically investigated flow matching for visual navigation or its integration with advanced temporal encoders, which is the focus of our work.

\subsection{State Space Models and Mamba}

State space models (SSMs)\cite{gu2021efficiently} introduced efficient continuous-time SSM parameterization, while Mamba~\cite{gu2024mamba} achieved a breakthrough with input-dependent selective mechanisms and hardware-aware parallel scanning. Mamba-2~\cite{dao2021transformers} further bridged SSMs and attention mechanisms theoretically, and bidirectional extensions such as Hydra~\cite{hwang2024hydra} and Vision Mamba~\cite{zhu2024vision} adapted the architecture for tasks requiring bidirectional context. While SSMs have been explored for robotic policy learning~\cite{cao2024mamba}, their application to temporal encoding in visual navigation remains largely unexplored. Our work is the first to systematically investigate Mamba as a temporal encoder combined with flow matching~\cite{lipman2023flow}, demonstrating its advantages over Transformer-based encoders, particularly for long observation histories.

\section{METHOD}
\begin{figure}[h!]
    \centering
    \includegraphics[width=0.48\textwidth]{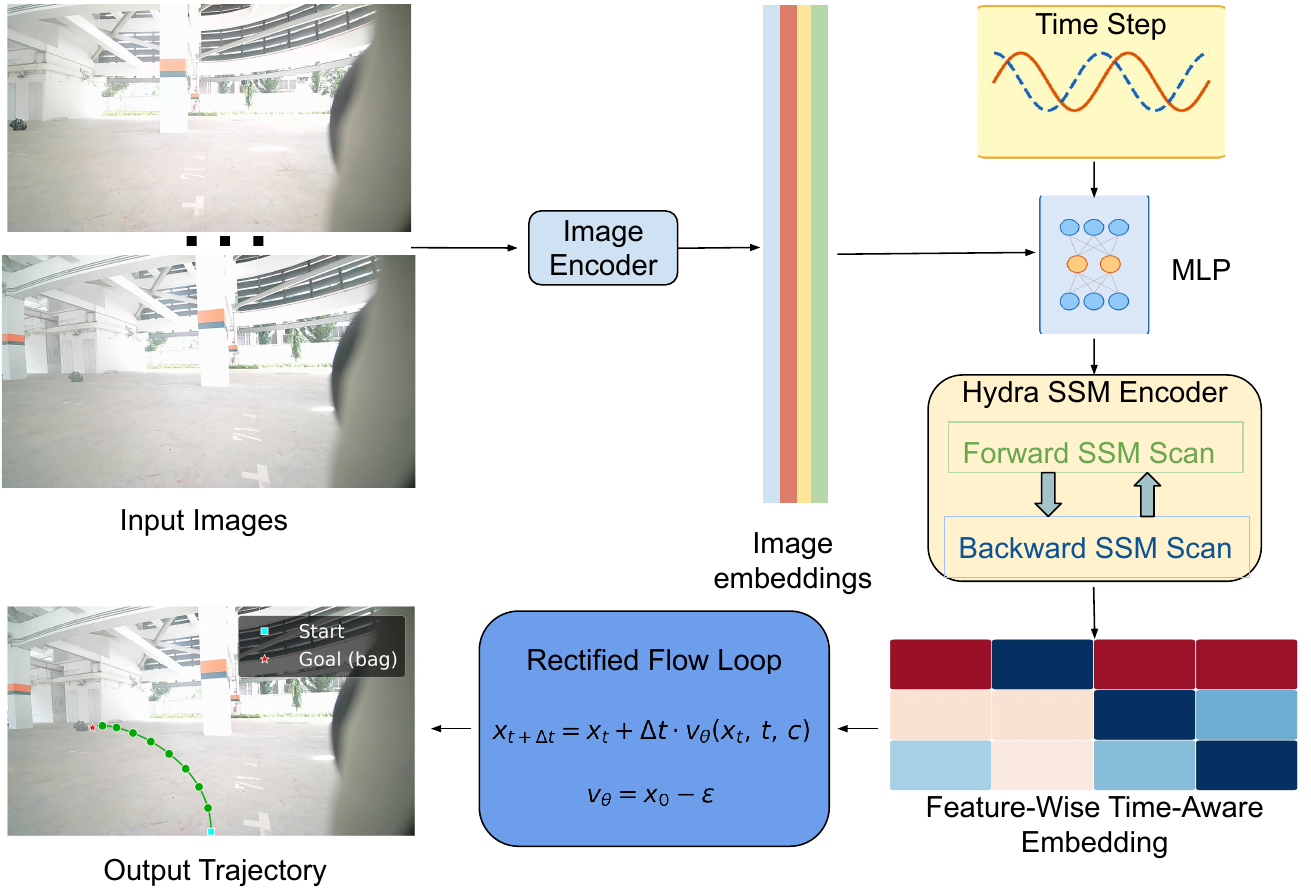}
    \vspace{-7mm}
    \caption{The pipeline of our architecture. The pipeline processes images through an image encoder to generate image embeddings, which are refined by a Hydra SSM Encoder performing bidirectional temporal scans. The temporally-enriched features are converted to time-injected embeddings through an MLP , which feed into a rectified flow loop to compute the final output trajectory.}
    \vspace{-4mm}
    \label{fig:pipeline}
\end{figure}

In this section, we present our proposed framework for efficient and accurate visual navigation. Our baseline architecture is built upon the NoMaD~\cite{sridhar2024nomad} model, and we introduce three principled improvements that collectively address the key limitations of existing navigation policies. 
Our framework consists of: 
(1) a rectified flow matching formulation that replaces the conventional DDPM denoising~\cite{song2020denoising} process to enable single-step inference; 
(2) a bidirectional Hydra~\cite{hwang2024hydra} encoder that replaces the transformer-based temporal encoder to more effectively capture long-range dependencies across multi-frame observations; 
and (3) a feature-wise conditioning mechanism that resolves the critical incompatibility between the Hydra encoder and flow matching, enabling the two improvements to work synergistically. An overview of the complete architecture is illustrated in Figure~\ref{fig:pipeline}.

\subsection{Rectified Flow Matching for Action Generation}
\label{sec:flow}
The NoMaD~\cite{sridhar2024nomad} baseline employs a DDPM-based~\cite{ho2020denoising} diffusion policy to generate waypoint action trajectories. During training, the forward diffusion process progressively corrupts a ground-truth action trajectory $a_0$ by adding Gaussian noise over $T$ discrete timesteps according to a predefined variance schedule $\{\beta_1, \ldots, \beta_T\}$:
\begin{equation}
    q(a_t \mid a_{t-1}) = \mathcal{N}(a_t; \sqrt{1 - \beta_t} \, a_{t-1}, \beta_t \mathbf{I}).
    \label{eq:ddpm_forward}
\end{equation}

At inference time, the model must iteratively reverse this process starting from pure Gaussian noise $a_T \sim \mathcal{N}(0, \mathbf{I})$, applying the learned denoising network for each of the $T$ steps to recover the clean action trajectory. NoMaD~\cite{sridhar2024nomad} requires $T = 10$ denoising steps, each involving a full forward pass through the UNet-based denoising network. 
To speed up the inference, we carefully replace the DDPM~\cite{ho2020denoising} formulation with rectified flow matching~\cite{liu2023flow}, which enables high-quality action trajectory generation in as few as a single integration step.

Flow matching~\cite{lipman2023flow} defines a continuous-time generative model by learning a velocity field $v_\theta$ that transports samples from a simple prior distribution $p_0$ (e.g., standard Gaussian) to the data distribution $p_1$. The transport is governed by an ODE:

\begin{equation}
    \frac{dx_t}{dt} = v_\theta(x_t, t), \quad t \in [0, 1],
    \label{eq:flow_ode}
\end{equation}

\noindent where $x_0 \sim p_0 = \mathcal{N}(0, \mathbf{I})$ is a noise sample and $x_1 \sim p_1$ is a data sample. The key insight of rectified flow~\cite{liu2023flow} is to define the interpolation path between noise and data as a simple \textit{straight line}:

\begin{equation}
    x_t = (1 - t) \, x_0 + t \, x_1, \quad t \in [0, 1],
    \label{eq:rectified_interpolation}
\end{equation}

\noindent where $x_0 \sim \mathcal{N}(0, \mathbf{I})$ is a noise sample and $x_1$ is a ground-truth action trajectory. The time derivative of this interpolation yields the constant velocity:

\begin{equation}
    \frac{dx_t}{dt} = x_1 - x_0,
    \label{eq:rectified_velocity}
\end{equation}

\noindent which is independent of $t$. The velocity prediction network $v_\theta$ is trained to regress this target velocity conditioned on the noisy sample $x_t$, the timestep $t$, and the context vector $c$:

\begin{equation}
    \mathcal{L}_{\text{flow}} = \mathbb{E}_{t \sim \mathcal{U}(0,1), \, x_0 \sim \mathcal{N}(0, \mathbf{I}), \, x_1 \sim p_{\text{data}}} \left\| v_\theta(x_t, t, c) - (x_1 - x_0) \right\|^2.
    \label{eq:flow_loss}
\end{equation}

Because the rectified flow formulation learns straight-line transport paths, the ODE in Eq.~\eqref{eq:flow_ode} can be approximated with a single Euler integration step at inference time:

\begin{equation}
    \hat{x}_1 = x_0 + v_\theta(x_0, t=0, c), \quad x_0 \sim \mathcal{N}(0, \mathbf{I}),
    \label{eq:single_step_inference}
\end{equation}

\noindent where $\hat{x}_1$ is the predicted action trajectory. This reduces the inference cost from $T = 10$ sequential forward passes in DDPM~\cite{ho2020denoising} to a single forward pass, yielding a decrease in latency. 

Compared to the DDPM~\cite{ho2020denoising} formulation, the flow matching approach offers several advantages for the navigation task: (1) the straight-line interpolation provides a more direct and stable training signal than the curved noising schedule of DDPM~\cite{ho2020denoising}; (2) the single-step inference eliminates the accumulation of errors across multiple denoising steps; and (3) the continuous-time formulation provides a natural framework for trading off between speed and quality by adjusting the number of Euler steps if needed.

\subsection{Hydra-Based Temporal Encoder}
\label{sec:mamba}
To capture input images' bidirectional dependency, we adopt a bidirectional Hydra architecture~\cite{hwang2024hydra} that processes the observation sequence in both forward and backward directions. 
The Hydra~\cite{hwang2024hydra} architecture is built upon the Structured State Space Sequence (S4) model~\cite{gu2021efficiently}, which maps a one-dimensional input sequence $x(t) \in \mathbb{R}$ to an output $y(t) \in \mathbb{R}$ through a latent state $h(t) \in \mathbb{R}^N$ via the following continuous-time linear ODE:
\begin{align}
    h(t) &= \mathbf{A} h(t) + \mathbf{B} x(t), \label{eq:ssm_continuous_h} \\
    y(t) &= \mathbf{C} h(t) + \mathbf{D} x(t), \label{eq:ssm_continuous_y}
\end{align}

\noindent where $\mathbf{A} \in \mathbb{R}^{N \times N}$ is the state transition matrix, $\mathbf{B} \in \mathbb{R}^{N \times 1}$ is the input projection matrix, $\mathbf{C} \in \mathbb{R}^{1 \times N}$ is the output projection matrix, and $\mathbf{D} \in \mathbb{R}$ is the skip connection parameter. 

Concretely, given the observation sequence $\{o_{t-K+1}, \ldots, o_t\}$, each frame is first encoded by a frozen ViT~\cite{dosovitskiy2020image} backbone to obtain per-frame embeddings $\{e_1, e_2, \ldots, e_K, e_g\}$. These embeddings are then processed by the Hydra~\cite{hwang2024hydra} encoder:
\begin{equation}
    \{z_1, z_2, \ldots, z_K, z_g\} = \text{Hydra}(\{e_1, e_2, \ldots, e_K, e_g\}),
    \label{eq:bimamba_encoding}
\end{equation}

\noindent followed by a learned pooling operation to produce the goal-conditioned context vector $c$.

This context vector $c$ encodes the full spatiotemporal observation history along with goal information, and serves as the conditioning signal for the downstream action generation policy. The linear complexity of the Hydra encoder ensures that increasing the context window size incurs only a proportional increase in computation, enabling the use of significantly longer observation histories compared to the transformer-based baseline.

\subsection{Feature-Wise Time-Aware Conditioning}
\label{sec:time_conditioning}

Having introduced rectified flow and the Hydra encoder as two individual
improvements, a natural question is whether they can be directly combined
to obtain cumulative benefits. Surprisingly, our experiments show that
the naive combination provides little additional gain and can even
underperform the Hydra-only variant. We therefore examine how the
rectified-flow state interacts with the temporally enriched visual
representation produced by Hydra.
Following Sec.~III-A, rectified flow constructs a linear interpolation
between a Gaussian sample $x_0 \sim \mathcal{N}(0,I)$ and a ground-truth
action trajectory $x_1$:
\begin{equation}
x_t = (1-t)x_0 + t x_1, \qquad t \in [0,1].
\label{eq:ftc_interpolation}
\end{equation}
The velocity network is trained with
\begin{equation}
\mathcal{L}_{\mathrm{FM}}
=
\mathbb{E}_{t,x_0,x_1}
\left[
\left\|
v_\theta(x_t,t,c) - (x_1-x_0)
\right\|_2^2
\right],
\label{eq:ftc_loss}
\end{equation}
where $c$ denotes the goal-conditioned temporal context produced by the
Hydra encoder.
Under the squared regression
objective, the optimal predictor can be expressed as
\begin{equation}
v^*(x,t,c)
=
\mathbb{E}
\left[
x_1-x_0
\mid
x_t=x,\; t,\; c
\right].
\label{eq:optimal_velocity}
\end{equation}
The conditional distribution of $(x_0,x_1)$ given the current state
$x_t=x$ generally changes with $t$. Consequently, the velocity field can
remain flow-state dependent even though the target associated with each
individual interpolation pair is constant. Explicit timestep conditioning
therefore remains useful for rectified-flow generation.
However, in the naive combination of rectified flow and Hydra, the timestep $t$ is
provided to the downstream velocity network, while the Hydra-derived
context $c$ remains unchanged for all interpolation states. Therefore,
the model must learn the interaction between the current flow state and
the temporally enriched visual context implicitly inside the velocity
network. We hypothesize that this weak coupling limits the effective use
of the rich temporal representation provided by Hydra, particularly when
the observation history becomes long.

To strengthen the interaction between the flow state and the temporal
visual representation, we introduce \emph{Feature-Wise Time-Aware
Conditioning} (FTC). Rather than using the same Hydra feature $c$ for all
$t$, FTC directly modulates the context according to the current flow
timestep. Given $t\in[0,1]$, we first compute a sinusoidal timestep
embedding
\begin{equation}
e(t)
=
[
\sin(t\omega_0), \cos(t\omega_0), \ldots,
\sin(t\omega_{d/2-1}), \cos(t\omega_{d/2-1})
],
\label{eq:time_embedding}
\end{equation}
where
\begin{equation}
\omega_k
=
\exp
\left(
-\frac{\ln 10000}{d/2-1}k
\right),
\qquad
k=0,\ldots,d/2-1,
\end{equation}
and $d$ is the context dimension. The timestep embedding is then projected
through a two-layer MLP with a SiLU activation:
\begin{equation}
h(t)
=
W_2 \sigma(W_1 e(t)+b_1)+b_2.
\label{eq:time_mlp}
\end{equation}
From $h(t)$, we predict feature-wise scale and shift parameters
\begin{equation}
[\gamma(t),\beta(t)]
=
W_3 h(t)+b_3,
\qquad
\gamma(t),\beta(t)\in\mathbb{R}^{d}.
\label{eq:gamma_beta}
\end{equation}
The Hydra context is then modulated as
\begin{equation}
\tilde{c}(t)
=
\left(1+\gamma(t)\right)\odot c+\beta(t),
\label{eq:ftc}
\end{equation}
where $\odot$ denotes element-wise multiplication. The resulting
flow-matching objective becomes
\begin{equation}
\mathcal{L}
=
\mathbb{E}_{t,x_0,x_1}
\left[
\left\|
v_\theta(x_t,t,\tilde{c}(t))
-
(x_1-x_0)
\right\|_2^2
\right].
\label{eq:ftc_objective}
\end{equation}

Compared with conventional timestep conditioning, FTC introduces an
additional direct pathway from the flow timestep to the visual
conditioning signal. Conceptually, the dependence of the velocity
prediction on $t$ can be decomposed as
\begin{equation}
\frac{\partial v_\theta}{\partial t}
=
\underbrace{
\frac{\partial v_\theta}{\partial e(t)}
\frac{\partial e(t)}{\partial t}
}_{\text{conventional timestep pathway}}
+
\underbrace{
\frac{\partial v_\theta}{\partial \tilde{c}}
\frac{\partial \tilde{c}}{\partial t}
}_{\text{feature-wise conditioning pathway}}.
\label{eq:time_gradient}
\end{equation}
The first pathway is already available when the timestep is provided
directly to the velocity network. FTC adds the second pathway, allowing
the flow timestep to directly control how the temporally aggregated
visual features are scaled and shifted. Thus, instead of requiring the
velocity network to learn the interaction between $t$ and $c$ entirely
implicitly, the proposed mechanism explicitly exposes this interaction
at the feature level.

\section{EXPERIMENTS}
We conduct extensive experiments to evaluate our proposed framework across multiple public benchmarks and a challenging internal dataset. We compare against state-of-the-art models ViNT~\cite{shah2023vint} and NoMaD~\cite{sridhar2024nomad} together with several ablation variants to validate the contribution of each proposed component. In addition to offline quantitative evaluation, we deploy our model on a physical quadruped robot to assess real-world navigation performance. 
In our experiments, F10 represents for increasing NoMad~\cite{sridhar2024nomad}'s input number of image frames from three to ten, which is a ablation study to verify the effectiveness of our architecture when inputting long image sequences; Ours (RF) represents for our model with rectified flow, Ours (Hydra) represents for our model with Hydra encoder, Ours (Na\"{i}ve) represents for the na\"{i}ve combination of RF and Hydra, and Ours (Full) represens for our full architecture.

\subsection{Datasets}
\label{sec:datasets}

We evaluate our framework on two widely used public benchmarks and one internally collected dataset, covering diverse environments and navigation scenarios. For fair comparison, we train separate models for each dataset across all our model variants and baseline methods from existing works.

\noindent\textbf{RECON}~\cite{shah2021rapid}. The RECON dataset contains navigation trajectories collected across a variety of outdoor environments including parks, trails, and urban sidewalks. The dataset features diverse visual conditions such as varying lighting, seasonal changes, and different terrain types. We select 150 representative trajectories for train and 20 trajectories for testing.

\noindent\textbf{TartanDrive}~\cite{triest2022tartandrive}. The TartanDrive dataset consists of off-road driving data collected on an all-terrain vehicle equipped with multiple sensors. The dataset features challenging unstructured environments with rough terrain, vegetation, and varying weather conditions. This dataset tests the model's ability to handle visually complex and geometrically challenging navigation scenarios. We select 150 representative trajectories for train and 20 trajectories for testing.

\noindent\textbf{Internal Quadruped Dataset.} To evaluate our framework in a more challenging real-world setting, we collect a custom dataset using the Deep Robotics X30 Pro quadruped platform. The robot is equipped with a front-facing RGB camera and records navigation trajectories across indoor office environments, outdoor paths, and semi-structured warehouse settings. Critically, our internal task is more demanding than standard point-goal navigation: the robot must not only navigate to the target location but also execute a \textit{circling motion} around the target upon arrival. 
This compound task requires the policy to generate smooth, continuous trajectories that testing the model's ability to produce diverse and temporally coherent action sequences. The dataset contains 120 navigation trajectories, split into 80\%/10\%/10\% for training, validation, and testing, respectively.

\subsection{Implementation Details}
\label{sec:implementation}

All models are trained using the AdamW optimizer with an initial learning rate of $1 \times 10^{-4}$, weight decay of $1 \times 10^{-2}$, and a cosine annealing learning rate schedule with a linear warmup of 500 steps. The batch size is set to 256 across all experiments. 
For the flow matching formulation, the timestep $t$ is uniformly sampled from $\mathcal{U}(0, 1)$ during training. 
For real-world deployment on the X30 Pro platform, the model runs on an onboard NVIDIA Jetson Orin module.
For experimental comparisons, we fine-tune ViNT~\cite{shah2023vint} and NoMad~\cite{sridhar2024nomad} using suggested hyper-parameters of their official code.

\subsection{Evaluation Metrics}
\label{sec:metrics}

We employ a comprehensive set of evaluation metrics to assess both offline prediction quality and real-world navigation performance.

\noindent\textbf{Waypoint Cosine Similarity (WCS).} We compute the cosine similarity between the predicted waypoint trajectory $\hat{a} = \{\hat{w}_1, \ldots, \hat{w}_H\}$ and the ground-truth trajectory $a = \{w_1, \ldots, w_H\}$ following NoMad~\cite{sridhar2024nomad}:
\begin{equation}
    \text{WCS} = \frac{1}{H} \sum_{i=1}^{H} \frac{\hat{w}_i \cdot w_i}{\|\hat{w}_i\| \, \|w_i\|},
    \label{eq:wcs}
\end{equation}
where $H$ is the trajectory horizon. Higher values indicate better directional alignment between predicted and ground-truth trajectories.

\begin{figure*}[h!]
    \centering
    \includegraphics[width=1\textwidth]{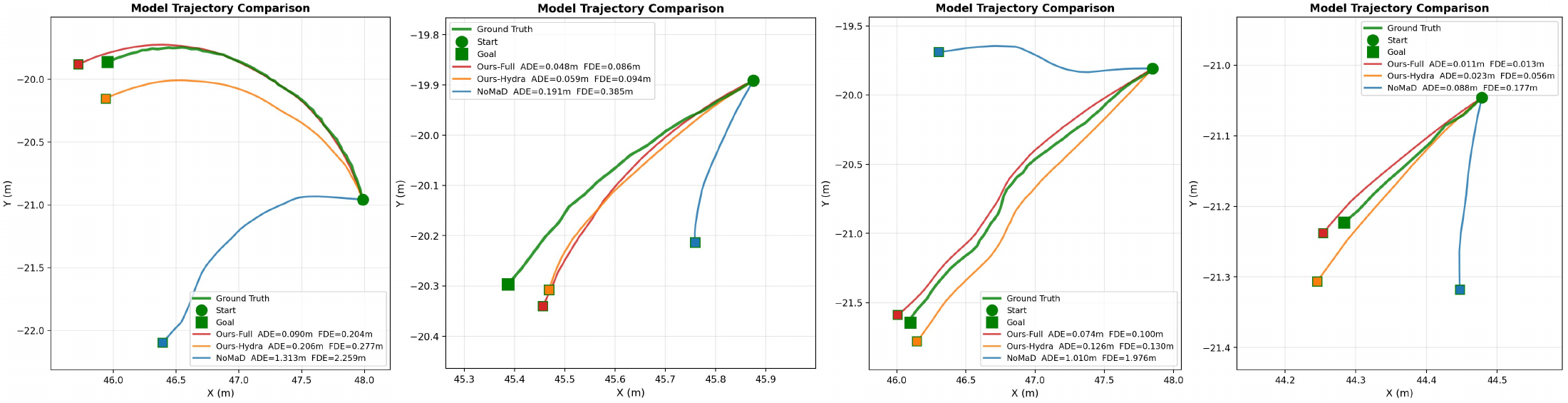}
    \vspace{-7mm}
    \caption{Qualitative comparisons on predicted trajectories from different models on our self-collected dataset.}
    \label{fig:our_data}
\end{figure*}

\begin{figure*}[h!]
    \centering
    \includegraphics[width=1\textwidth]{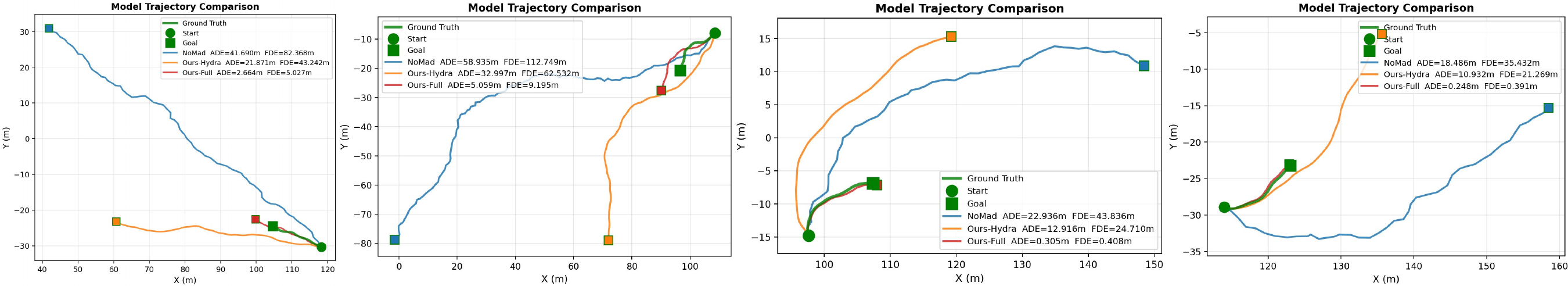}
    \vspace{-7mm}
    \caption{Qualitative comparisons on predicted trajectories from different models on the RECON~\cite{shah2021rapid} dataset.}
    \vspace{-4mm}
    \label{fig:recon}
\end{figure*}

\noindent\textbf{Average Displacement Error (ADE).} We report the mean Euclidean distance between all predicted waypoints and their corresponding ground-truth waypoints: \begin{equation} \text{ADE} = \frac{1}{H} \sum_{i=1}^{H} |\hat{w}_i - w_i|_2. \label{eq:ade} \end{equation} This metric measures the overall trajectory accuracy by averaging the displacement error across all $H$ waypoints, capturing how closely the predicted path aligns with the ground-truth trajectory as a whole. In Figures.~\ref{fig:our_data} \& \ref{fig:recon}, ADE is reported as the the mean of all the frames in world coordinates, while ADE in Tabs~\ref{tab:recon}, \ref{tab:our_data}, \ref{tab:certis}, \ref{tab:randomness} \& \ref{tab:tartan} is reported as the mean of predicted waypoints in robot coordinates. The same for FDE.

\noindent\textbf{Final Displacement Error (FDE).} We report the Euclidean distance between the final predicted waypoint and the final ground-truth waypoint:
\begin{equation}
    \text{FDE} = \|\hat{w}_H - w_H\|_2.
    \label{eq:fde}
\end{equation}
This metric specifically measures the accuracy of the predicted goal-reaching position.

\noindent\textbf{Navigation Success Rate (SR).} For real-world experiments on the X30 Pro platform, we define a navigation trial as successful if the robot reaches around a 1-meter radius of the target location and successfully finish the circular motion around the target object.

\noindent\textbf{Inference Time.} 
We report the average time to generate one action trajectory, measured on an
  NVIDIA RTX PRO 6000 GPU in Table~\ref{tab:time}. Inference latency is dominated
  by the number of sequential passes through the action (denoising) network rather
  than by the visual encoder, since the encoder runs only once per prediction
  whereas the denoising network is invoked repeatedly. NoMaD~\cite{sridhar2024nomad}
  requires ten sequential U-Net passes to iteratively denoise the trajectory,
  while our rectified-flow formulation recovers the trajectory in a single Euler
  integration step. Adopting the bidirectional Hydra encoder does add per-call cost, which is shown in Ours~(Hydra) at $98.7$~ms under the same ten-step sampling, but
  this cost is incurred only once, and Mamba's linear-time recurrence keeps its
  growth modest even as the number of input frames increases. Consequently,
  combining single-step rectified flow with the Hydra encoder (Ours~(Full)) reduces
  the latency to $68.4$~ms, which is not only below the multi-step Hydra variant
  but also faster than the NoMaD baseline's $79.9$~ms~\cite{sridhar2024nomad},
  despite our model processing substantially longer observation sequences.

\subsection{Quantitative Comparisons}
\label{sec:quantitative}
\begin{table}[t]
    \centering
    \caption{Quantitative comparison of navigation performance on the Recon dataset~\cite{shah2021rapid}. We report WCS $\uparrow$, ADE $\downarrow$ and FDE $\downarrow$. }
    \label{tab:recon}
    \vspace{-4mm}
    \begin{tabular}{l ccc}
    \hline
        \textbf{Method} & \textbf{WCS $\uparrow$} & \textbf{ADE $\downarrow$} & \textbf{FDE $\downarrow$}\\
        \hline
        ViNT~\cite{shah2023vint}       & 0.8785 & 2.6486m & 4.8585m\\
        NoMaD~\cite{sridhar2024nomad}  & 0.9024 & 2.6336m & 4.6816m\\
        NoMaD~\cite{sridhar2024nomad} (F10)  &0.9090  & 2.6359m & 4.7013m\\
        Ours (RF)          & 0.9093 & 2.0675m & 4.1776m  \\
        Ours (Hydra)                   & 0.9191  & 1.1651m & 1.4249m\\
        Ours (Na\"{i}ve)               & 0.9178 & 1.5301m & 1.9512m  \\
        Ours (Full)                     & \textbf{0.9550} & \textbf{0.0872m} & \textbf{0.1170m}\\
        
        \hline
        \vspace{-5mm}
    \end{tabular}
\end{table}

\begin{table}[t]
    \centering
    \caption{Quantitative comparison of navigation performance on our collected dataset. We report WCS $\uparrow$, ADE $\downarrow$ and FDE $\downarrow$. }
    \label{tab:our_data}
    \vspace{-4mm}
    \begin{tabular}{l  ccc}
    \hline
        \textbf{Method} & \textbf{WCS $\uparrow$} & \textbf{ADE $\downarrow$} & \textbf{FDE $\downarrow$} \\
        \hline
        ViNT~\cite{shah2023vint}       & 0.8512 & 0.5740m & 1.1212m \\
        NoMaD~\cite{sridhar2024nomad}  & 0.8672 & 0.5280m & 0.9272m \\
        NoMaD~\cite{sridhar2024nomad} (F10)  &0.9012  &0.0618m  & 0.1058m \\
        Ours (RF)          & 0.9004 & 0.0640m & 0.1034m \\
        Ours (Hydra)                   & 0.9195 & 0.0590m & 0.0994m \\
        Ours (Na\"{i}ve)               & 0.9105 & 0.0610m & 0.1004m \\
        Ours (Full)                     & \textbf{0.9411} & \textbf{0.0178m} & \textbf{0.0233m}\\
        \hline
        \vspace{-10mm}
    \end{tabular}
\end{table}

\begin{table}[t]
    \centering
    \caption{Randomness evaluation with mean and standard deviation for our full model.}
    \label{tab:randomness}
    \vspace{-3mm}
    \resizebox{\columnwidth}{!}{%
    \begin{tabular}{l cccc}
    \hline
        \textbf{Datasets} & \textbf{WCS} & \textbf{ADE} & \textbf{FDE} \\
        \hline
        RECON~\cite{shah2021rapid}  & $0.9550 \pm 3.11\times 10^{-3}$ & $0.0872 \pm 9.95\times 10^{-3}$m & $0.1170 \pm 1.71\times 10^{-3}$m \\
        Ours   & $0.9411 \pm 6.77\times 10^{-3}$ & $0.0178 \pm 3.81\times 10^{-3}$m & $0.0233 \pm 7.18\times 10^{-3}$m \\
        \hline
    \end{tabular}
    }
    \vspace{-3mm}
\end{table}
\begin{table}[t]
    \centering
    \caption{Quantitative comparison of navigation performance on the TartanDrive dataset~\cite{triest2022tartandrive}. We report WCS $\uparrow$, ADE $\downarrow$ and FDE $\downarrow$.}
    \label{tab:tartan}
    \vspace{-3mm}
    \begin{tabular}{l  cccc}
    \hline
        \textbf{TartanDrive~\cite{triest2022tartandrive}} & \textbf{WCS $\uparrow$} & \textbf{ADE $\downarrow$} & \textbf{FDE $\downarrow$} \\
        \hline
        NoMaD~\cite{sridhar2024nomad}  & \textbf{0.8054} & 0.5951m & 1.0573m \\
        Ours (Full)   & 0.7953 & \textbf{0.1955m} & \textbf{0.2593m}\\
        \hline
        \vspace{-10mm}
    \end{tabular}
\end{table}
\begin{table}[t]
    \centering
    \caption{Inference time comparison among different models on the RECON~\cite{shah2021rapid} dataset.}
    \label{tab:time}
    \vspace{-4mm}
    \begin{tabular}{l  cccc}
    \hline
        \textbf{Method} & \textbf{Nomad~\cite{sridhar2024nomad}} & \textbf{Ours (Hydra)} & \textbf{Ours (Full)}\\
        \hline
        Time       &79.9ms  &98.7ms  & \textbf{68.4ms}\\
        \hline
        \vspace{-7mm}
    \end{tabular}
\end{table}
\begin{table}[t]
    \centering
    \caption{Quantitative comparison on the real-world deployment on the navigation accuracy (SR). FDT represents for the time percentage that robot can detect the full target during the circular motion.}
    \label{tab:certis}
    \vspace{-3mm}
    \resizebox{\columnwidth}{!}{
    \begin{tabular}{l  cccc}
    \hline
        \textbf{Method} & ViNT~\cite{shah2023vint} & NoMaD~\cite{sridhar2024nomad} & Ours (Hydra) & Ours (Full) \\
        \hline
        \textbf{SR $\uparrow$}  &86.7\% &86.7\% &86.7\% &\textbf{100\%} \\
        \textbf{FDT $\uparrow$} &75.8\% &76.1\% &79.9\% &\textbf{80.2\%} \\
        \hline
        \vspace{-10mm}
    \end{tabular}
    }
\end{table}

Tables~\ref{tab:recon} and~\ref{tab:our_data} summarize the quantitative
results on the public RECON~\cite{shah2021rapid} dataset and our internal dataset, respectively.
We also report the randomness evaluation in Table~\ref{tab:randomness}, demonstrating the stability of our model across two datasets.

On the public RECON dataset (Table~\ref{tab:recon}), our full model
with feature-wise timestep conditioning achieves the best
performance across all three metrics.
Compared to the strongest baseline
NoMaD~\cite{sridhar2024nomad}, Ours-Full improves WCS by 5.83\% while reducing ADE and FDE by 96.69\% and 97.50\%, respectively. 
Among our ablation variants, replacing only the diffusion policy with rectified flow (Ours-RF) yields a modest WCS improvement over NoMaD and provides gains in ADE and FDE, indicating that one-step generation alone can slightly improve the performance at the same time increases the inference speed. Adopting the bidirectional Hydra encoder (Ours-Hydra) brings more
substantial improvements, reducing ADE to 1.1651\,m and FDE to 1.4249\,m,
which highlights the benefit of temporal modeling across observation frames.
However, na\"{i}vely combining rectified flow with the Hydra encoder
(Ours-Na\"{i}ve) degrades performance relative to Ours-Hydra. 
This result
suggests that simply providing the flow timestep to the downstream velocity
network is insufficient to fully exploit the temporally enriched Hydra
representation.
In contrast, Ours-Full directly adapts the context according
to the current flow state.

The results on our internal dataset (Table~\ref{tab:our_data}) exhibit
consistent trends. Ours-Full again achieves the best overall performance
with a WCS of 0.9411, an ADE of 0.0184\,m, and an FDE of 0.0240\,m,
substantially outperforming existing works. Notably, all four of our variants including Ours-RF
surpass the baselines by a considerable margin on this dataset, 
suggesting the generalization of the single-step inference from rectified flow formulation.
Nevertheless, the same
pattern that the na\"{i}ve combination underperforming
Ours-Hydra in both ADE and FDE persists, while Ours-Full reduces these errors further. This consistent
improvement across both datasets validates the effectiveness and
generalizability of the proposed feature-wise timestep conditioning
mechanism in bridging the rectified flow policy and the bidirectional
Hydra encoder. 
Similar trends are also observed in TartanDrive~\cite{triest2022tartandrive} dataset as shown in Table~\ref{tab:tartan}.

Furthermore, we also conduct an experiment to increase the number of input frames from three to ten for NoMad (NoMad F10) to compare with the results of Our-Hydra, we can see that even though simply increase the input number can obtain better performances, but they are inferior compared to ours Hydra variant. 

\subsection{Qualitative Results}
\label{sec:qualitative}
To further illustrate the effectiveness of our approach, we visualize
different predicted trajectories alongside the ground truth on both
our self-collected dataset and the public RECON~\cite{shah2021rapid} dataset in
Figures~\ref{fig:our_data} and~\ref{fig:recon}, respectively.

As shown in Figure~\ref{fig:our_data}. NoMaD exhibits the largest
deviation from the ground truth, and Ours-Hydra achieves a superior performances, with its trajectory visually
closer to the ground truth yet still exhibiting minor deviations at
turning points. Our full model produces the most faithful
trajectory, achieving the lowest ADE and FDE,
closely tracing the ground-truth path throughout the entire navigation
sequence. 
Similarly, as shown in Figure~\ref{fig:recon}, the baseline NoMaD produces
trajectories that deviate substantially from the ground truth. Introducing the
bidirectional Hydra encoder (Ours-Hydra) further improves trajectory
alignment, which
suggests that capturing temporal dependencies across observation frames
contributes meaningfully to directional accuracy. However, a large gap
relative to the ground truth still persists. In
contrast, our full model with feature-wise timestep conditioning
produces a trajectory that nearly overlaps with the ground
truth, achieving a dramatically lower ADE and FDE. 

Notably, while there are differences in environment scale and complexity for the two datasets with the absolute error magnitudes differ between
them,
the relative ranking among methods remains similar. This cross-dataset consistency reinforces the generalizability of our approach and confirms that the feature-wise timestep conditioning mechanism is essential for unlocking the full potential of combining rectified flow with bidirectional temporal encoding. 

\begin{figure}[h!]
    \centering
    \includegraphics[width=0.48\textwidth]{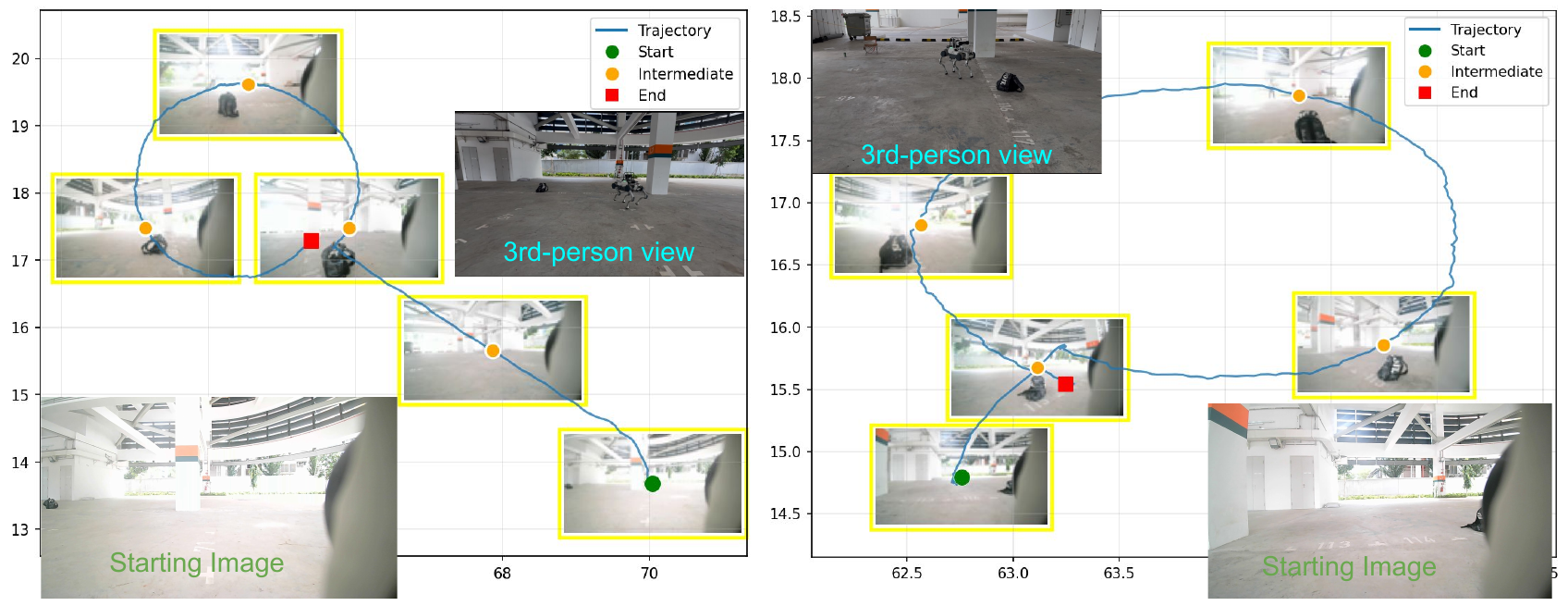}
    \vspace{-8mm}
    \caption{Visualization of the detection-guided navigation with two phases, the 1st one is normal motion and the 2nd one is circular motion around a target object.}
    \vspace{-5mm}
    \label{fig:certis_testing}
\end{figure}

\subsection{Real-World Deployment with Navigation and Circling}
\label{sec:realworld}

To evaluate the practical effectiveness of our framework beyond offline metrics, we deploy all methods on a physical robot platform tasked with navigating toward a designated target and subsequently performing a circular inspection motion around it. We train two separate models for the two phases, namely navigation toward the target and circular inspection, and combine them during deployment. Both models share the same architecture but are trained on different data corresponding to their respective phases. We report two metrics: navigation success rate (SR) and full detection time (FDT), defined as the percentage of the circular inspection phase during which the robot maintains continuous full visibility of the target. Results are summarized in Table~\ref{tab:certis}.

We designed a comprehensive set of experiments across five distinct starting points, each tested three times for a total of fifteen trials. The robot's camera is equipped with a 130-degree field of view (FOV), which informed the selection of starting positions. 
1) The first starting point was positioned directly in front of the target object at 5 meters, serving as a baseline for straight-line navigation accuracy. 
2) The second and third starting points were placed at the leftmost and rightmost extremes of the camera's FOV at the 5-meter range, which is designed to test whether the robot could accurately reach the object from the farthest perceivable positions, thereby verifying that cumulative navigation error remained sufficiently small over long distances. 
3) The fourth and fifth starting points were positioned at a very short vertical distance of 1.5 meters from the robot, but at the maximum horizontal offset where the camera could just barely detect the object at the edges of its FOV. These near-but-lateral positions were specifically chosen to test whether the robot could execute a sufficiently large turning angle within a very short distance to orient itself toward and reach the target. 
The examples are shown in Figure~\ref{fig:certis_testing}. 

The quantitative comparisons among different models are shown in Table~\ref{tab:certis}. Existing works or our model variants suffer from the inability to simultaneously achieve a sufficiently large turning angle in a short period of time and maintain sufficiently small accumulated error over long distances, and they failed 2 times out of 15 trials.
The performance gap in SR (86.7\% → 100\%) demonstrates that our model provides a meaningful improvement in navigation reliability, particularly in challenging starting positions.
Most notably, Ours-Full achieves the highest FDT of 80.2\%
among all methods.
This metric is
particularly meaningful as it directly reflects the quality of
the predicted circular trajectory during the inspection phase:
a higher FDT indicates that the robot maintains a consistent
and appropriate radius around the target, keeping it fully
within the sensor's field of view throughout the circular motion.
This capability is essential for security inspection scenarios,
where the robot must perform a complete circumnavigation of a
potentially hazardous target to gather comprehensive visual
information from all angles. 
We attached a video showing the robot navigating toward the bag from both lateral directions and executing a full circular motion around the target.

\section{CONCLUSIONS}
In this work, we presented a unified framework for efficient vision-based
robotic navigation comprising three contributions: a rectified flow
formulation that reduces multi-step denoising to a single inference step,
a bidirectional Mamba-based visual encoder for capturing temporal
dependencies across observation frames, and a feature-wise timestep
conditioning mechanism that bridges the two by modulating encoded features
via learned scale and shift parameters derived from the flow timestep.
Our key finding is that na\"{i}vely combining rectified flow with the
temporal encoder degrades performance, and our
feature-wise conditioning addresses this limitation by allowing the visual
context to adapt as a function of the flow timestep during training.
Experiments across multiple datasets and real-world deployment demonstrate
consistent improvements over the corresponding baselines.










\bibliographystyle{IEEEtran}
\bibliography{IEEEabrv,references}


\end{document}